\documentclass[10pt,twocolumn,letterpaper]{article}

\usepackage[pagenumbers]{cvpr} % To force page numbers, e.g. for an arXiv version

\definecolor{cvprblue}{rgb}{0.21,0.49,0.74}
\usepackage[pagebackref,breaklinks,colorlinks,allcolors=cvprblue]{hyperref}

\usepackage{multirow}
\usepackage[most]{tcolorbox}
\usepackage{float}
\newtcolorbox[]{mybox}[3][]{
    arc=1mm,
    fonttitle=\bfseries,
    colbacktitle=NavyBlue!10,
    coltitle=blue!50!black,
    enhanced,
    colframe=blue!50!black,
    colback=white!10,
    overlay={},
    overlay={},
    title=#2 \thetcbcounter,#1,breakable
}
\definecolor{darkgreen}{RGB}{0,150,0}
\definecolor{darkred}{RGB}{200,0,0}
\definecolor{darkblue}{RGB}{0,0,200}
\usepackage{dblfloatfix}

\def\paperID{11803} % *** Enter the Paper ID here
\def\confName{CVPR}
\def\confYear{2025}

\title{MarkushGrapher: Joint Visual and Textual Recognition of Markush Structures}

\author{
  Lucas Morin\textsuperscript{1, 2} \quad Valéry Weber\textsuperscript{1} \quad Ahmed Nassar\textsuperscript{1} \quad Gerhard Ingmar Meijer\textsuperscript{1} \\ \quad Luc Van Gool \textsuperscript{2, 3} \quad Yawei Li \textsuperscript{2} \quad Peter Staar\textsuperscript{1}\\
  \textsuperscript{1}IBM Research \quad \textsuperscript{2}ETH Zurich \quad \textsuperscript{3}INSAIT \\
  {\tt\small \{lum, vwe, ahn, inm, taa\}@zurich.ibm.com} \quad {\tt\small \{yawei.li, vangool\}@vision.ee.ethz.ch}  
}

\begin{document}
\maketitle

% Lucas Comment
% Clearances paper + code + data + Llama

\begin{abstract}
The automated analysis of chemical literature holds promise to accelerate discovery in fields such as material science and drug development. In particular, search capabilities for chemical structures and Markush structures (chemical structure templates) within patent documents are valuable, \eg, for prior-art search. Advancements have been made in the automatic extraction of chemical structures from text and images, yet the Markush structures remain largely unexplored due to their complex multi-modal nature. In this work, we present MarkushGrapher, a multi-modal approach for recognizing Markush structures in documents. Our method jointly encodes text, image, and layout information through a Vision-Text-Layout encoder and an Optical Chemical Structure Recognition vision encoder. These representations are merged and used to auto-regressively generate a sequential graph representation of the Markush structure along with a table defining its variable groups. To overcome the lack of real-world training data, we propose a synthetic data generation pipeline that produces a wide range of realistic Markush structures. Additionally, we present M2S, the first annotated benchmark of real-world Markush structures, to advance research on this challenging task. Extensive experiments demonstrate that our approach outperforms state-of-the-art chemistry-specific and general-purpose vision-language models in most evaluation settings. Code, models, and datasets will be available \footnote{https://github.com/DS4SD/MarkushGrapher}.
\end{abstract}

%\vspace{-2mm}

\section{Introduction}

\begin{figure}\vspace{0mm}
    \centering
    \includegraphics*[trim={3.25cm 0.75cm 3.25cm 0.5cm}, clip, width=0.5\textwidth]{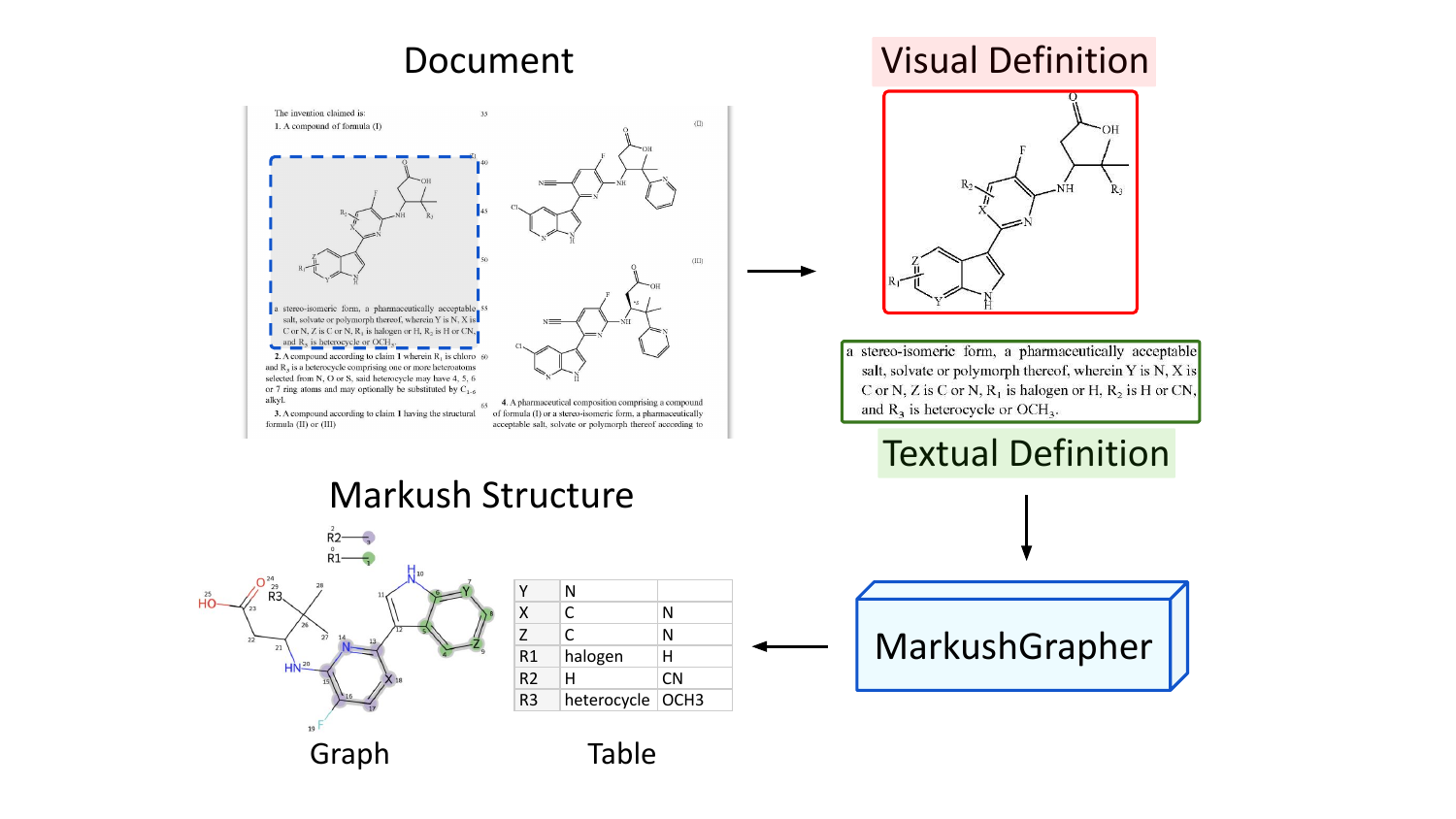}\vspace{-2mm}
    \caption{\textbf{MarkushGrapher} extracts Markush structures from documents using their visual and textual definitions.}
    \label{fig:introduction}\vspace{-4mm}
\end{figure}

Document understanding plays a critical role in accelerating discovery in chemistry. For example, databases \cite{Morin2024} can be built to offer a unified view of the current knowledge in documents. 
Such databases can substantially facilitate and accelerate Research \& Development in fields ranging from drug discovery to material science \cite{Pyzer-Knapp2025}. 
In documents, molecules can be referred to using textual names or depictions of their molecular structures. Over the past years, substantial advances have been made to extract molecules from text using chemical name entity recognition \cite{zhai-etal-2019-improving,doi:10.1021/acs.jcim.6b00207,Korvigo2018}, as well as, from images through chemical structure image segmentation \cite{Rajan2021,Zhou2023,Tang2024} and chemical structure recognition \cite{Morin_2023_ICCV,Qian2023,Rajan2023}. More recently, advances in multi-modal document understanding have opened up opportunities to extract chemical structure templates, the so-called Markush structures \cite{SIMMONS2003195}. These widely used representations are compact descriptions of families of related molecules. A Markush structure consists of a combination of an image and a text element: the image defines the Markush backbone, which contains atoms, bonds, and variable regions, while the accompanying text element specifies the molecular substituents that can replace those variable regions.
The variable regions can be variable groups (commonly referred to as R-groups), frequency variation indicators, and positional variation indicators \cite{doi:10.1021/ci00002a009}. A visualization of these variable regions is provided in the suppl. materials. 
These structures are particularly valuable for prior-art search, freedom-to-operate assessment, or landscape analysis in patent documents \cite{SIMMONS2003195}. Yet, only a very few databases containing Markush structures exist (CAS MARPAT \cite{Ebe1991}, CAS DWPIM \cite{doi:10.1021/ci00066a008}, and WIPO Patentscope), all of which are proprietary and manually created. Multi-modal Markush Structure Recognition (MMSR) is a critical task towards automatically creating and scaling Markush structure databases.

MMSR presents several substantial challenges that currently limit the effectiveness of automatic recognition methods. First, the immense diversity of Markush backbones, resulting from the combinatorial explosion of atom-and-bond associations, makes good generalization particularly difficult for data-driven deep learning models. Second, Markush backbones can be depicted in a wide range of styles and conventions. Third, Markush textual descriptions employ various formats for representing substituents, such as non-structured names, molecular string identifiers, and chemical formulas. In more complex cases, the descriptions may contain interdependent variable definitions. Finally, the lack of real-world training datasets adds another layer of complexity to applying deep learning methods in this field. 

Initial approaches focused on subparts of the MMSR task. Among them, some methods can recognize a subset of Markush backbone images \cite{Rajan2023,chen2024molnextrgeneralizeddeeplearning,Qian2023}. From the cropped image, they apply a deep network to directly output a string of characters identifying the Markush backbone, in the form of a Simplified Molecular Input Line Entry System (SMILES) \cite{SMILES} or a sequential graph representation. These models extend the task of recognizing molecule images, known as Optical Chemical Structure Recognition (OCSR), by generalizing the recognition of abbreviations to variable groups. However, variable groups are only one specific feature of Markush structure, and these methods do not handle the full complexity of Markush backbones, such as frequency variation indicators. 
Other works are designed to analyze variable groups definitions from textual descriptions only. Such methods are formulated as a classification task, where labels, and definitions in the textual description are classified. 
These text-only and image-only methods are limited as they do not exploit the dependencies between variables' textual definitions and their context in the Markush backbone.

Recently, some toolkits have combined textual and visual information for chemical document understanding \cite{doi:10.1021/ci800449t,doi:10.1021/acs.jcim.4c00572}. However, they still rely on separate models to handle both modalities. At the same time, document understanding has seen the emergence of multi-modal models \cite{huang2022layoutlmv3pretrainingdocumentai,tang2023unifyingvisiontextlayout}. They perform a wide range of tasks on documents, using a unified architecture which takes visual and textual inputs, and outputs images and texts as well. Such document understanding models were trained on documents in chemistry \cite{wei2024generalocrtheoryocr20}, and more specifically, to answer if a query molecule is covered by the Markush structures in a document \cite{cai2024unismartuniversalsciencemultimodal}. In this scenario, the model must implicitly recognize multi-modal Markush structures. However, using this method to search through large sets of documents is not practical, because it would require to run the model on every document for each user query.

In this work, we introduce MarkushGrapher for Multi-modal Markush Structure Recognition, illustrated in \autoref{fig:introduction}. The encoder-decoder model takes as input an image of a Markush structure and the OCR cells of all text depicted in the image, and outputs a text sequence representing the structure. This output sequence contains two parts, one represents the graph of the Markush backbone image, and one represents a table of substituents which instantiates the variables in the backbone. 
The image and OCR cells are jointly encoded using two encoders, a Vision-Text-Layout (VTL) encoder and an Optical Chemical Structure Recognition (OCSR) encoder. These encodings are then concatenated and used by a text decoder to autoregressively generate a Markush sequential representation.

To further aid the research in the community, we introduce M2S, a benchmark dataset of annotated real Markush structures. 
We also release our pipeline for generating synthetic Markush structures used to train MarkushGrapher. Comprehensive experiments are performed on three benchmarks: MarkushGrapher-Synthetic, M2S and USPTO-Markush. Our model outperforms general-purpose as well as chemistry-specific document understanding models.
In summary:
\begin{itemize}
    \item We develop MarkushGrapher, which combines a VTL encoder and an OCSR encoder to recognize multi-modal Markush structures. Our method achieves state-of-the-art performance for joint recognition of visual and textual definitions of Markush structures. 
    \item We introduce M2S, a benchmark of manually-annotated multi-modal Markush structures from patent documents.
    \item We build a synthetic data generation pipeline to generate a wide variety of images and accompanying text descriptions of Markush structures.
\end{itemize}

\section{Related Work}

\begin{figure*}[t]
    \centering \vspace{0mm}
    \includegraphics*[trim={0.4cm 4.1cm 0.4cm 4.3cm}, clip, width=0.95\textwidth]{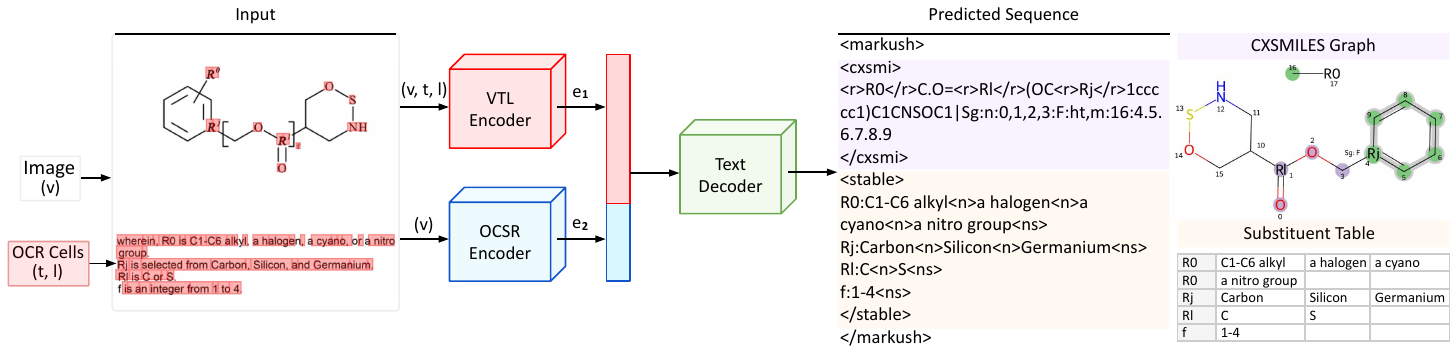} \vspace{-2mm}
    \caption{\textbf{Markush structure recognition architecture}. MarkushGrapher jointly encodes the input image and its text with a VTL encoder (blue) and an OCSR encoder (red). The VTL output ($e_{1}$) and the OCSR output ($e_{2}$) are concatenated. Finally, this joint encoding is processed with a text decoder to predict a sequential representation of the Markush backbone (purple) and its substituent table (orange).}
    \label{fig:architecture}\vspace{-3mm}
\end{figure*}

\noindent\textbf{Markush Structure Recognition.} Existing work on Markush structure recognition typically treats vision and text understanding as separate tasks. On one hand, some models focus exclusively on visual understanding of Markush structures. These recognition methods are often adapted from OCSR techniques \cite{oldenhof2023weakly, Oldenhof_2024_CVPR}, which targets standard molecular structures. They commonly employ a vision encoder paired with an autoregressive text decoder, as seen in DECIMER \cite{Rajan2023}, MolScribe \cite{Qian2023}, MolNextTR \cite{chen2024molnextrgeneralizeddeeplearning}, and MPOCSR \cite{Lin2024}. By only covering variable groups represented as abbreviations, these methods omit most Markush structure features. While Image2SMILES \cite{https://doi.org/10.1002/cmtd.202100069} incorporates Markush images with position variation indicators, it remains limited in scope. In contrast, our method tackles all Markush structure features: variable groups, frequency and positional variation indicators. %Besides, alternative methods have been developed to classify Markush structures visually \cite{Jurriaans2023OneSY}.
On the other hand, some approaches exclusively target the recognition of the textual Markush structure component. This task can be formulated as a classification task, and approached, for example, using BioBERT \cite{DCM34463} or hand-written grammars \cite{Lowe2015}. Additionally, several toolkits aim to combine text-only and image-only models for document understanding in chemistry, including CliDE \cite{doi:10.1021/ci800449t}, for chemical structure recognition, and OpenChemIE \cite{doi:10.1021/acs.jcim.4c00572} for reaction recognition. Specifically, Beard and Cole \cite{Beard2020}, Haupt \cite{Haupt2009}, and Jie, \textit{et al.}, \cite{10.1093/bib/bbac461} proposed toolkits for reconstructing multi-modal Markush structures from text-only and image-only models. However, for MMSR, it is preferred to jointly analyze visual and textual modalities to effectively exploit the dependencies between textual variable definitions and their context in the Markush backbone (see \autoref{section:ablation}). While Jie, \textit{et al.}, generate a structural library from Markush structures, our method extracts a compact representation. This is crucial for scalability as, for example, Wagner, \textit{et al.}, estimated that a single patent EP0810209B1 already contains about $5\times10^{16}$ molecules \cite{WAGNER2022104597}.

\noindent\textbf{Document Understanding.} Document understanding involves extracting a document's content and structure from a page image input \cite{auer2024doclingtechnicalreport, Mishra_2024}. Recent advancements integrate textual and visual information through unified models, addressing multiple tasks including document classification, layout segmentation, question answering, information extraction, key-value extraction, table structure recognition, and chart question answering \cite{huang2022layoutlmv3pretrainingdocumentai, hu2024mplugdocowl15unifiedstructure, wei2024generalocrtheoryocr20, ye2023ureader, kim2022ocr, nassar2025smoldoclingultracompactvisionlanguagemodel}. Certain models, such as UDOP \cite{tang2023unifyingvisiontextlayout}, rely on an external OCR input and introduce a technique for merging these OCR text tokens together with the image patches containing them. %These unified approaches have implications for the chemistry domain as well. For instance, $\alpha$-extractor \cite{Xiong2024} employs a unified architecture for molecular structure image segmentation and recognition.
More specifically, initial progress has been made in extending general document understanding models for the recognition of Markush structures. Uni-SMART \cite{cai2024unismartuniversalsciencemultimodal} determines whether a query molecule can be found in a document, directly as a molecule, or covered by one of the Markush structures in the document. However, this approach does not enable the explicit extraction of Markush structures, rendering it impractical for large-scale applications.
%Nevertheless, training such models involves the inefficient process of collecting question-answer pairs about whether a document covers specific molecules.
%\vspace{-5mm}
\section{MarkushGrapher}

We introduce MarkushGrapher, a Vision-Text-Layout transformer for MMSR. The model architecture is illustrated in \autoref{fig:architecture}.
First, the input is encoded using a unified Vision-Text-Layout (VTL) encoder and a Optical Chemical Structure Recognition (OCSR) vision encoder. Second, a text decoder autoregressively predicts from this encoding an optimized Markush sequential representation. Finally, this output is parsed to resolve the graph of the Markush structure backbone and its table of variable groups.

\subsection{Architecture}

MarkushGrapher takes as input three components: an image, the text content in the image, and the positions of this text provided by bounding boxes. For MMSR, the text, visual, and layout modalities are interdependent. On the Markush backbone, text cells represent the atoms, abbreviations, and variable groups, which are positioned via the bounding boxes, while the image defines the bonds and other non-textual objects. Both are needed together to correctly predict structures. Additionally, the text definition of variable groups gives prior information to the neighboring regions of these groups within the image (see \autoref{section:ablation}). Besides that, relying on an external OCR, instead of training MarkushGrapher with image-only input has multiple advantages. First, an OCR model can be trained on a wide variety of datasets with fine-grained supervision making it highly robust with various typography. Second, using an OCR is needed to recognize complex abbreviations \cite{Morin_2023_ICCV} whose diversity can not be fully covered in training sets. Third, recent work on OCSR shows that relying on localized image features greatly reduces the required number of training samples and improves performance \cite{Qian2023, NEURIPS2022_ada36dfe}. By providing atom positions as input, the model can easily identify such local features. The three input modalities - image, text and position - are then encoded using two separate encoders. 

On the one hand, a Vision-Text-Layout encoder jointly encodes the three modalities following the UDOP's approach \cite{tang2023unifyingvisiontextlayout}. Let $e_{1}$ be the output encoding of the $\text{VTL}$ encoder,
\begin{equation}
e_{1}(v, t, l) = \text{VTL}(v, t, l) \,.
\end{equation}

In this setup, the image ($v$) is first divided into patches. Image patches and text ($t$) tokens are independently projected to obtain visual and textual embeddings. Then, the layout ($l$) information is used to merge both. 
For each text cell, the embedding of the image patch containing its center is selected. Then, the corresponding visual and textual embeddings are summed. Next, a sequence is formed by appending the joint visual-textual embeddings, followed by the remaining visual embeddings that do not overlap with any text region.
%The layout information is implicitly encoded through this merging. 
The resulting sequence is used as input to a standard transformer encoder. More details on the joint text-vision-layout encoding technique can be found in the UDOP publication \cite{tang2023unifyingvisiontextlayout}. 

On the other hand, the image is encoded by an OCSR vision encoder. Let $e_{2}$ be its output encoding,
\begin{equation}
e_{2}(v) = \text{OCSR}(v) \,.
\end{equation}
We use a pre-trained vision encoder implemented as an OCSR transformer. The image is encoded and then projected using a Multi-Layer Perceptron ($\text{MLP}$). Because of the similarities between the Markush structures and chemical structure images, this approach benefits from the robust features learned by OCSR models trained on larger training sets and on real data.
This projected OCSR encoding is finally concatenated with the VTL output,
\begin{figure}[]
\centering \vspace{0mm}
    \includegraphics[trim={1.75cm 5cm 1.75cm 4cm}, clip, width=0.475\textwidth]{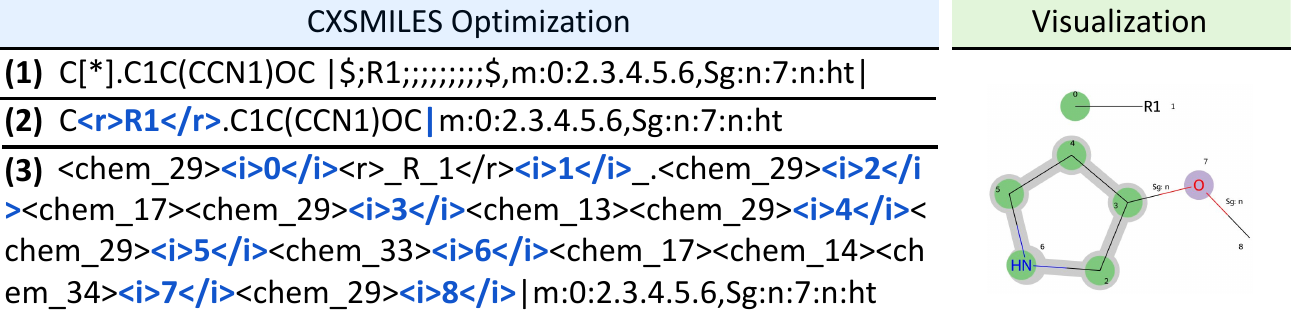}\vspace{-1mm}
    \caption{\textbf{Optimized CXSMILES format.} The figure presents the steps of the CXSMILES optimization. The CXSMILES (\textbf{1}) is first compacted by moving variable groups in the SMILES sequence and removing unnecessary characters (\textbf{2}). Then, the indices of atoms are appended after each atom (between $<i>$ and $</i>$ tokens) and the sequence is encoded using a specific vocabulary for atoms and bonds ($<chem>$ tokens) (\textbf{3}).}
    \label{fig:format}\vspace{-3mm}
\end{figure}
\begin{equation} %\boldsymbol
e(v,t,l) = e_{1}(v,t,l) \oplus \text{MLP}(e_{2}(v)) \,.
\end{equation}
%
% Decoder
From this joint encoding, a text decoder iteratively predicts a sequence representing a Markush structure. More details in \autoref{section:Implementation-Details}.

\begin{figure*}[t]    
    \centering\vspace{0mm}
    \includegraphics*[trim={0cm 4.25cm 0cm 4.25cm}, clip, width=\textwidth]{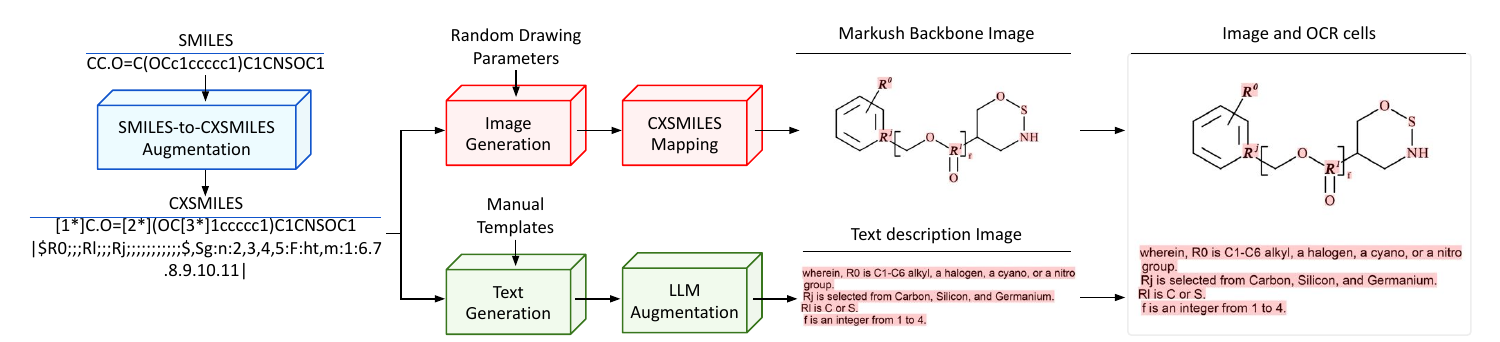}\vspace{-1mm}
    \caption{\textbf{Synthetic training data generation}. The figure presents the pipeline for generating synthetic training samples. First, a molecule is sampled from PubChem and augmented to create a CXSMILES. Second, the CXSMILES is used to jointly generate a image of the Markush backbone and its OCR cells (red), and generate an image of a text description and its OCR cells (green) . Finally, images are collated to create a training sample.}
    \label{fig:data}\vspace{-3mm}
\end{figure*}

\subsection{Optimized Markush Structure Representation}

Next, we describe the Markush structure representation in MarkushGrapher, shown in \autoref{fig:format}.
Markush structure backbones are commonly represented using a string identifier named Chemaxon Extended SMILES (CXSMILES) \cite{cxsmiles}. The CXSMILES format extends the standard SMILES format, used to store atoms and bonds of molecular structures. CXSMILES strings begin with a standard SMILES, and append an extension block, which allows to represent variable groups, position variation indicators (denoted as `m' sections in the CXSMILES), and frequency variation indicators (denoted `Sg' section in the CXSMILES). Substituents for each variable group can also be stored. Illustrations are provided in the suppl. materials. 

To facilitate the auto-regressive prediction of the CXSMILES, we optimized its format (see Fig. 2), making it more compact. Uninformative characters (spaces and ending `$|$') are removed. To shrink the required attention window, we move the variable groups from the extension section to the SMILES by representing them as special atoms. Also, we add to each atom its SMILES index. As the extension section uses the indices to identify atoms, this extra information explicitly links atoms in the SMILES and the extension section. Moreover, we arrange the atom indices in the `m' sections in ascending order to maintain a consistent format. Finally, we use a specific vocabulary for atoms and bonds rather than encoding them as standard characters. Instead of storing substituents directly in the CXSMILES a substituent table is appended.
It enables to store free-text substituents and to compact the definition by grouping together variable groups that have the same substituents, as well as compressing lists of integers defining frequency variation variables.

\section{Datasets}

\subsection{Synthetic Training Data Generation}
\label{section:MarkushGrapher-Synthetic}

To the best of our knowledge, no training dataset of annotated Markush structures is available. Thus, we develop a synthetic generation pipeline, described in \autoref{fig:data}, to generate a broad diversity of visual and textual Markush representations. Generating synthetic Markush structures is a challenge as it requires aligning multi-modal data \cite{10.1093/bib/bbac461}. 

Our dataset is created using molecules represented by SMILES selected from the PubChem database \cite{10.1093/nar/gky1033}. To increase the variety of molecular structures, we sample SMILES based on substructure diversity. Using the RDKit \cite{RDKit} library, these SMILES are augmented to create artificial CXSMILES. (Cf. suppl. material for details.)

Next, we render these CXSMILES using the Chemistry Development Kit (CDK) \cite{Willighagen2017}.
Drawing parameters are randomly selected to increase diversity. These include font, bond width, spacing of bonds and atoms, index display as subscripts or superscripts, explicit or implicit carbon display, aromatic cycles with circles, and atom number display.
To extract the OCR cells from the synthetic images, we develop a parser for the SVG images generated with CDK. It yields the bounding boxes of text in the image. These boxes are mapped to their corresponding text, by aligning the SVG file with an associated MOL file \cite{MolFile}, which contains the molecular structures. %The same parser could be particularly useful to extract atom-level annotations to train an OCSR model \cite{doi:10.1021/ci00007a012}.

Furthermore, we generate a text description defining variable groups depicted in the image. For this description to be chemically correct and compatible with the CXSMILES, valence constraints must be respected. The descriptions are generated using multiple templates and mappings. These templates describe the structure of the sentence and are manually extracted from patent documents. They consist of sentences defining variable groups, lists of variable groups, or frequency variation labels. The sentences contain lists of variable groups, lists of substituents, and lists of integers, which are also defined using multiple templates. Additionally, some templates are used to append additional information at the beginning or the end of the description, at the beginning or the end of each item definition, or to add noise information to the description.
Then, these templates are instantiated based on mappings, which are lists of manually selected substituents from patent documents, common abbreviations represented by SMILES or names, molecular substructures represented by SMILES or names, and atoms represented by chemical symbols or names. 
Together with the description, we generate the substituent table. 

Finally, we use a LLM to paraphrase a fraction of these descriptions. The substituents table is provided as context to avoid changing any semantic information in the description, but only the writing style. This wide variety of generated descriptions helps generalization. 
Details on the data generation pipeline are available in the suppl. material.

\subsection{The M2S Dataset}
\label{section:M2S}

There is no public benchmark of annotated real data. Collecting annotations is tedious, hence Markush structure databases usually restrict their textual annotations to a set of keywords \cite{doi:10.1021/ci00031a010}. In our case, we chose that the substituents table can contain free-text to cover a large variety of cases.

We introduce M2S, a benchmark of annotated Markush structures from patent documents. The images are crops of Markush structure backbone images and their textual descriptions. It contains 103 images selected from patents published by the US Patent and Trademark Office (USPTO), European Patent Office (EPO) and World Intellectual Property Organization (WIPO). They are manually annotated with Markush structure backbones, stored as CXSMILES, OCR cells, and substituent tables. The images have a selection bias towards Markush structures in the claims, because these are the most relevant for patents. The Markush structure backbones were annotated on Chemaxon Marvin JS \cite{Marvin}, and the OCR cells were annotated using LabelStudio \cite{LabelStudio}.
See suppl. material for further details.

\section{Experiments}

This section describes experiments with multiple Multi-modal Markush Structure Recognition benchmarks. 

\subsection{Implementation Details}
\label{section:Implementation-Details}

\begin{table*}[t]\vspace{0mm}
\centering
\caption{\textbf{Comparison of our method with existing MMSR models.} Evaluation on synthetic (MarkushGrapher-Synthetic) and real data (M2S, USPTO-Markush) benchmarks. Exact match (EM) evaluates the percentage of perfect predictions. Tanimoto score (T) and F1-score (F1) evaluate the similarity between the prediction and ground-truth with a percentage between 0 (most dissimilar) and 100 (most similar).}
\label{tab:results}
\resizebox{0.95\linewidth}{!}{
\begin{tabular}{lccccccccccccccc}
\hline
\textbf{Methods} &  & \multicolumn{5}{c}{MarkushGrapher-Synthetic (1000)} &  & \multicolumn{5}{c}{M2S (103)} &  & \multicolumn{2}{c}{USPTO-Markush (74)} \\ \cline{3-7} \cline{9-13} \cline{15-16} 
 &  & \multicolumn{2}{c}{CXSMILES} & \multicolumn{2}{c}{Table} & \multicolumn{1}{c}{Markush} &  & \multicolumn{2}{c}{CXSMILES} & \multicolumn{2}{c}{Table} & \multicolumn{1}{c}{Markush} &  & \multicolumn{2}{c}{CXSMILES} \\
 &  & \multicolumn{1}{c}{EM} & \multicolumn{1}{c}{T} & \multicolumn{1}{c}{EM} & \multicolumn{1}{c}{F1} & \multicolumn{1}{c}{EM} &  & \multicolumn{1}{c}{EM} & \multicolumn{1}{c}{T} & \multicolumn{1}{c}{EM} & \multicolumn{1}{c}{F1} & \multicolumn{1}{c}{EM} &  & \multicolumn{1}{c}{EM} & \multicolumn{1}{c}{T} \\ \hline
\textbf{Image-only} &  &  &  &  &  &  &  &  &  &  &  &  &  &  &  \\
DECIMER \cite{Rajan2023} &  & 7 & 35 & - & - & - &  & 3 & 25 & - & - & - &  & 7 & 40 \\
MolScribe \cite{Qian2023} &  & 11 & 47 & - & - & - &  & 21 & 73 & - & - & - &  & 7 & \textbf{97} \\ \hline
\textbf{Multi-modal} &  &  &  &  &  &  &  &  &  &  &  &  &  &  &  \\
Pixtral-12B-2409 \cite{agrawal2024pixtral12b} &  & 0 & 3 & 9 & 35 & 0 &  & 0 & 3 & 7 & 27 & 0 &  & - & - \\
Llama-3.2-11B-Vision-Instruct &  & 0 & 2 & 3 & 7 & 0 &  & 0 & 1 & 1 & 11 & 0 &  & - & - \\
GPT-4o \cite{openai2024gpt4technicalreport} &  & 0 & 6 & 10 & 46 & 0 &  & 4 & 11 & 19 & 49 & 0 &  & - & - \\
Uni-SMART \cite{cai2024unismartuniversalsciencemultimodal} &  & - & - & - & - & - &  & 0 & 46 & 0 & 1 & 0 &  & - & - \\
\textbf{MarkushGrapher (Ours)} &  & \textbf{65} & \textbf{96} & \textbf{84} & \textbf{96} & \textbf{57} &  & \textbf{38} & \textbf{76} & \textbf{29} & \textbf{65} & \textbf{10} &  & \textbf{32} & 74 \\ \hline
\end{tabular}} \vspace{-2mm}
\end{table*}

%The implementation is done with PyTorch 2.2.2 and CUDA 12.1. 
The Vision-Text-Layout encoder and text decoder use a T5-large encoder-decoder architecture \cite{10.5555/3455716.3455856}. The OCSR encoder is the vision encoder of MolScribe \cite{Qian2023}. It is frozen during training. Overall, the model has 831M parameters (744M trainable).
It is trained on 210,000 synthetic images, presented in \autoref{section:MarkushGrapher-Synthetic}. We train for 10 epochs on a NVIDIA H100 GPU using ADAM with a learning rate of 5e-4, 100 warmup steps, a batch size of 10 and a weight decay of 1e-3.
The LLM used to augment 10 percent of synthetic text descriptions is Mistral-7B-Instruct-v0.3 \cite{jiang2024mixtralexperts}.
(Cf. suppl. material for details.)

\subsection{Evaluation Datasets and Metrics}

\textbf{Datasets.} For a comparison with SOTA, our method is evaluated on the benchmarks MarkushGrapher-Synthetic, M2S and USPTO-Markush. We also use a subset of the SciAssess benchmark \cite{cai2024sciassessbenchmarkingllmproficiency} for qualitative evaluation. M2S is described in \autoref{section:M2S}. MarkushGrapher-Synthetic is a set of 1000 synthetic Markush structures generated using the pipeline described in \autoref{section:MarkushGrapher-Synthetic}. The images are sampled such that overall, each Markush features (R-groups, 'm' and 'Sg' sections) is represented evenly.
SciAssess \cite{cai2024sciassessbenchmarkingllmproficiency} is a benchmark to assess LLMs in scientific literature analysis, covering topics such as biology, chemistry, material science, and medicine. In the chemistry domain, it evaluates if a molecule is contained in a document either, as a molecular structure image, or as a part of one of its Markush structures. The set contains 50 question-answer pairs drawn from 14 documents. As this task is not precisely aligned with our setup, we manually retrieve Markush structure backbone images from these documents and use them for qualitative evaluation.
USPTO-Markush is a dataset containing 75 Markush structure backbone images from USPTO patent documents \cite{USPTO}. Its images are annotated with OCR cells and CXSMILES and a large proportion includes indicators for positional and frequency variations.

\noindent\textbf{Metrics.} We introduce metrics to evaluate MMSR in two tasks: image recognition and substituent table recognition. For image recognition, we rely on CXSMILES exact match and Tanimoto similarity score \cite{tanimoto1958elementary}. The CXSMILES exact match measures the percentage of perfectly recognized CXSMILES. A match is exact if two conditions are met: (1) disregarding Markush features, the predicted SMILES matches the ground truth according to InChIKey \cite{InChI} equality, and (2) variable groups, as well as the `m' and `Sg' sections, are correctly represented. As to the Tanimoto score, it measures the structural similarity between two chemical compounds. In our case, we remove Markush features, encode structures using the RDKit DayLight fingerprint \cite{2011DaylightTM}, and compute a similarity score by comparing the bit vectors of these fingerprints using the Tanimoto coefficient. To compute the CXSMILES exact match and Tanimoto score, the stereo-chemistry is ignored.
For substituent table recognition, we measure the exact match and the F1-score similarity. The exact match measures the proportion of perfectly recognized tables, \ie, all variable groups and substituents are correct. To compute the F1-score, we first determine recall and precision by averaging the percentage of correctly retrieved (for recall) and correctly predicted (for precision) substituents per variable group, and then aggregate these averages across all variable groups. %It is worth to note that these text metrics consider as incorrect any semantically equivalent prediction, for instance "Nitrogen" instead of the annotated symbol "N". 
Finally, we report the Markush structure exact match, which requires both the CXSMILES exact match and substituent table exact match to be correct. 

\begin{figure*}[t]
\vspace{0mm}
\centering%
\includegraphics[trim={11cm 1.5cm 11cm 1.25cm}, clip, width=0.97\textwidth]{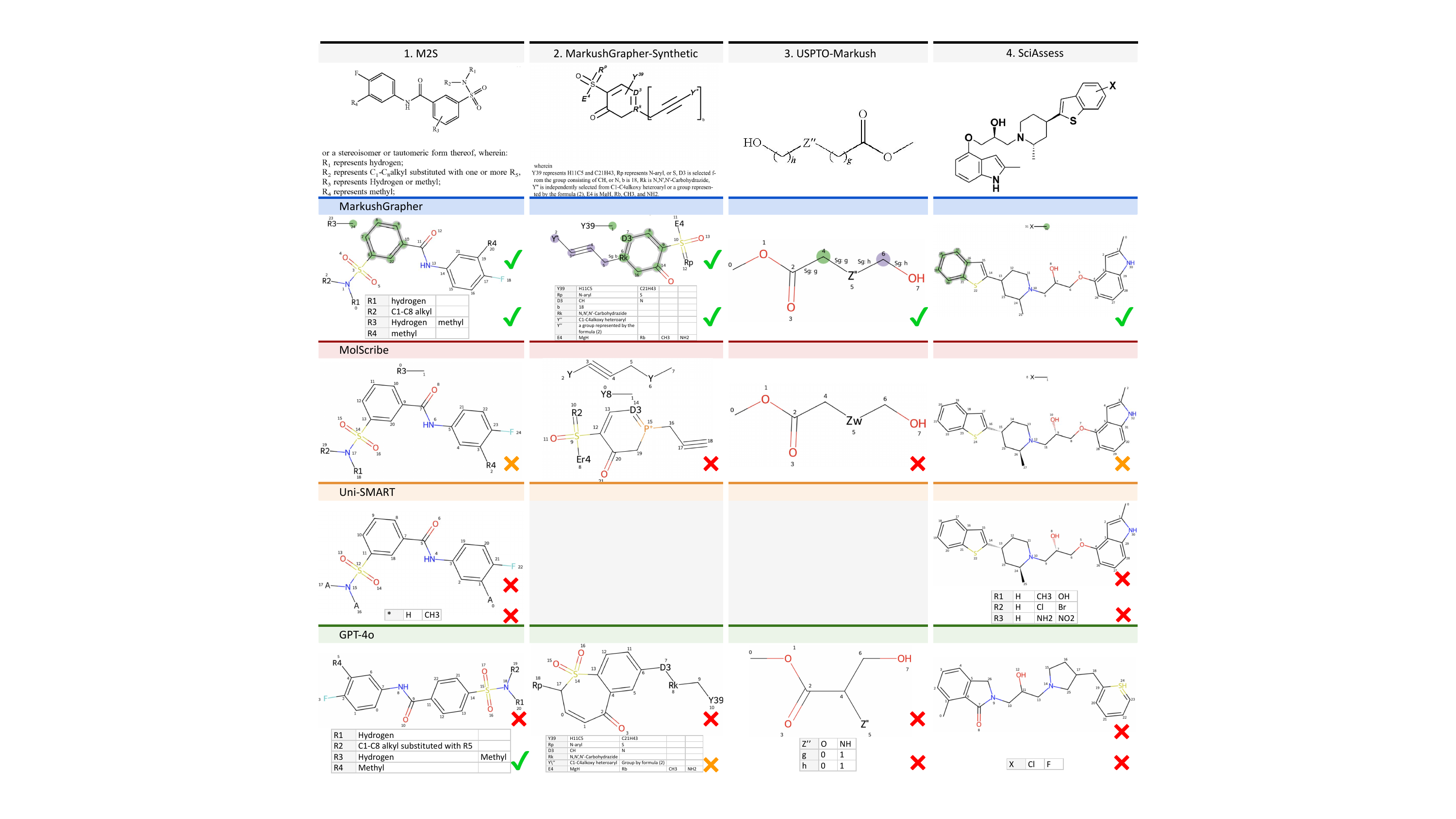}\vspace{-2.5mm}
\caption{\textbf{Qualitative comparison.} Examples of predictions are shown for the different MMSR models on real-world data (M2S, USPTO-Markush, SciAssess) and on synthetic data (MarkushGrapher-Synthetic).}
\label{fig:comparison}\vspace{-1mm}
\end{figure*}

\subsection{State-of-the-art Comparison}

\subsubsection{Multi-modal Markush Structure Recognition}

\autoref{tab:results} compares the MMSR methods for both visual and textual recognition of Markush structures across synthetic (MarkushGrapher-Synthetic) and real-world (M2S, USPTO-Markush) benchmarks. 
Our approach demonstrates superior performance to image-only, chemistry-specific models on most datasets. Notably, MarkushGrapher surpasses MolScribe in CXSMILES exact match accuracy on M2S (38\% against 21\%), USPTO-Markush (32\% against 7\%), and MarkushGrapher-Synthetic (65\% against 11\%). This substantial performance advantage highlights that DECIMER and MolScribe capture only a limited subset of Markush features, specifically R-groups represented as abbreviations.
Interestingly, MolScribe achieves a higher CXSMILES Tanimoto score on the USPTO-Markush dataset. This score, which disregards Markush-specific structural features, highlights MolScribe's robustness in identifying standard chemical structures. This observation inspired the integration of the OCSR encoder into MarkushGrapher. 

Our method substantially outperforms general-purpose VLMs in CXSMILES recognition. VLMs fail to recognize Markush structures. Although Uni-SMART is trained to verify if a molecule is covered within a document containing Markush structures, it cannot actually identify these structures {\em per se}, a simpler task by comparison. MarkushGrapher also surpasses Uni-SMART in terms of the CXSMILES Tanimoto score. Furthermore, MarkushGrapher outperforms VLMs in table recognition, achieving higher exact match (29\% vs. 19\%) and F1-score (65\% vs. 49\%). It is worth noticing that the table exact match and F1-score metrics may be disadvantageous for language models, as they often paraphrase content into semantically equivalent terms, which would be counted as errors; nonetheless, these metrics remain the most appropriate.

Overall, our multi-modal approach allows MarkushGrapher to outperform other methods on both the recognition of image and text in Markush structures.

\subsection{Markush Features Analysis}

\autoref{tab:analysis} shows the performance of MarkushGrapher vs. image-only MMSR models for the different features of Markush structures. 
MarkushGrapher has a clearly higher CXSMILES exact match performance, largely because DECIMER and MolScribe can not generate predictions for the `m' and `Sg' sections. While MarkushGrapher beats MolScribe at recognizing R-groups in the M2S dataset, it fails to do so in the USPTO-Markush dataset. This is probably due to the images of the M2S benchmark containing text descriptions, whereas USPTO-Markush consists solely of Markush backbone images.
The performance of MarkushGrapher is notably stronger than the other models on USPTO-Markush, with its higher proportion of images with `Sg' and `m' sections. Additionally, MarkushGrapher is more effective at recognizing R-groups and `m' sections compared to `Sg' sections. Diversifying the frequency variation indicators in synthetic examples could further improve the recognition of `Sg' sections.

\begin{table}[t]\vspace{-3mm}
\centering
\caption{\textbf{Comparison of Markush features performances.} Impact of the Markush features on the recognition performances. The percentage of correctly recognized R-groups (R), `m' sections (m), `Sg' sections (Sg), and CXSMILES (EM) are reported.}\vspace{-3mm}
\label{tab:analysis}
\resizebox{0.97\linewidth}{!}{
\begin{tabular}{llcccclcccc}
\hline
\multirow{2}{*}{\textbf{Methods}} &  & \multicolumn{4}{c}{M2S} &  & \multicolumn{4}{c}{USPTO-Markush} \\ \cline{3-6} \cline{8-11} 
 &  & R & m & Sg & EM &  & R & m & Sg & EM \\ \hline
DECIMER \cite{Rajan2023} &  & 12 & 0 & 0 & 3 &  & 34 & 0 & 0 & 7 \\
MolScribe \cite{Qian2023} &  & 61 & 0 & 0 & 21 &  & \textbf{76} & 0 & 0 & 7 \\
\textbf{MarkushGrapher (Ours)} &  & \textbf{75} & \textbf{76} & \textbf{31} & \textbf{38} &  & 69 & \textbf{67} & \textbf{18} & \textbf{32} \\ \hline
\end{tabular}} \vspace{-4mm}
\end{table}

\subsection{Qualitative Evaluation}

Next, we conduct a qualitative evaluation of MarkushGrapher vs. SOTA methods. \autoref{fig:comparison} shows examples of predicted molecules for images from various benchmark datasets. MarkushGrapher accurately recognizes complex Markush image features such as position variation indicators (\autoref{fig:comparison}, column 1, 2 and 4) and frequency variation indicators (\autoref{fig:comparison}, column 2 and 3). 
Contrary to Uni-SMART and GPT-4o our model does not predict an incorrect table when the input image does not contain any table (\autoref{fig:comparison}, column 3 and 4). 
Even when Markush features are disregarded in the evaluation, MarkushGrapher sometimes outperforms MolScribe (\autoref{fig:comparison}, column 3), likely because MolScribe struggles with text descriptions or the presence of Markush features within the images.
In addition, MarkushGrapher can handle the prediction of long tables which often confuse other models. (Cf. suppl. material for more details.)

\subsection{Ablation Study}
\label{section:ablation}

\begin{figure}[b]
\vspace{-3mm}
\centering%
    \includegraphics[trim={3.5cm 0.5cm 3.5cm 0.5cm}, clip, width=0.43\textwidth]{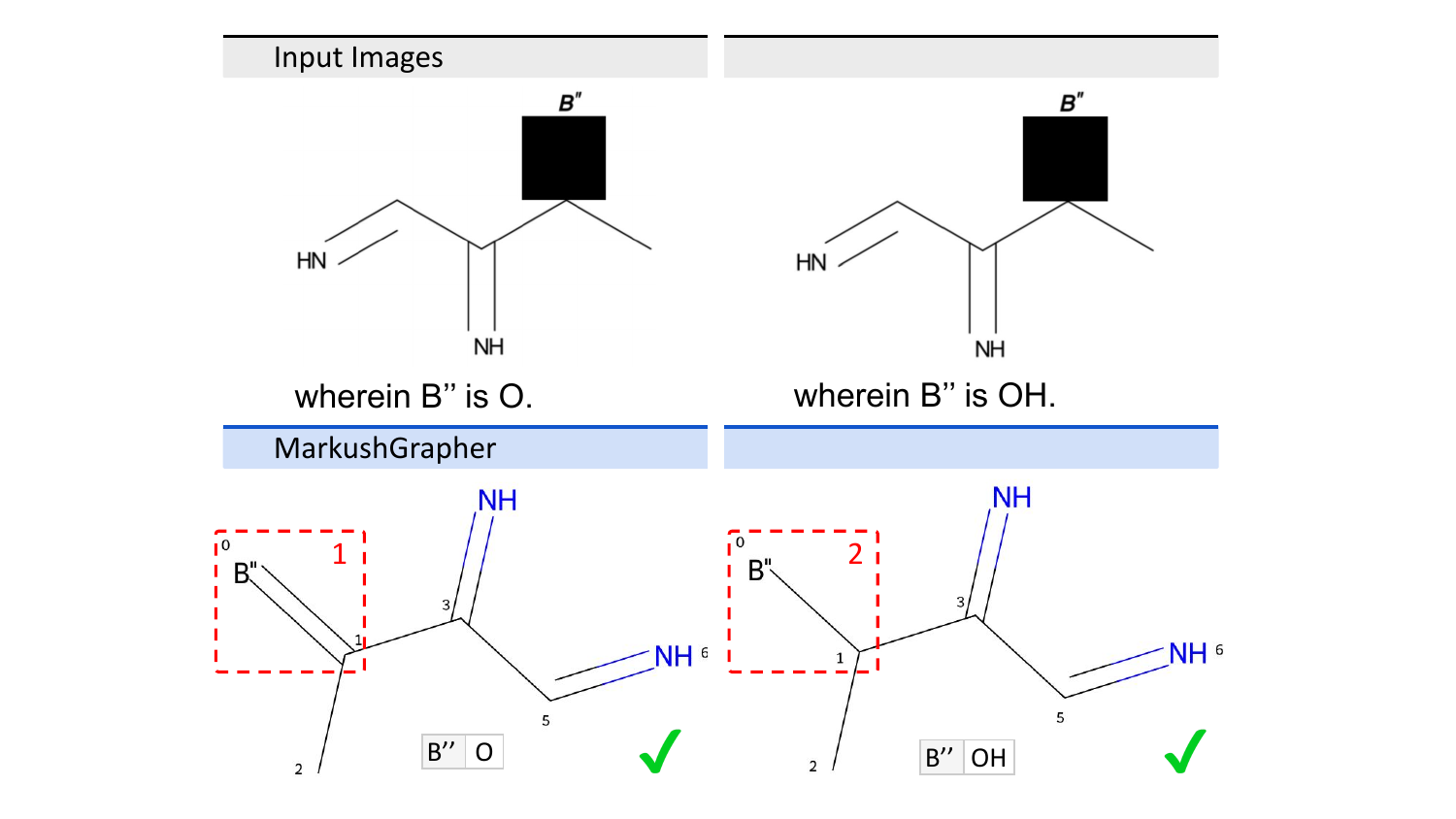}\vspace{-3mm}
    \caption{\textbf{Cross-modality understanding.} A bond is masked in the input image (black patch). For an oxygen atom O, MarkushGrapher predicts a double bond (red box 1). For an alcohol group OH, MarkushGrapher predicts a single bond (red box 2).}
    \label{fig:cross_modalities}\vspace{0mm}
\end{figure}

\noindent\textbf{MarkushGrapher encoder.} \autoref{tab:encoder} shows the impact of the MarkushGrapher encoder. It can be observed that using the OCSR encoder, in addition, to the VTL encoder give an improvement of performances on the USPTO-Markush dataset. 
We also compare two fusion alternatives. Early fusion (EF) corresponds to concatenating the OCSR encodings with the input of the VTL encoder, and late fusion (LF) corresponds to concatenating the OCSR encodings with the output of the VTL encoder. Late fusion yields better results.

\noindent\textbf{Optimized Markush structure representation.} \autoref{tab:representation} highlights the impact of Markush structure representation. Removing the R-group compression or the atom indexing both lower performances on M2S and USPTO-Markush. The indexing is crucial for USPTO-Markush, which contains more CXSMILES with `m' or `Sg' sections. 
 
\noindent\textbf{Modalities dependencies.} MarkushGrapher exploits dependencies between the text and image definitions of Markush structures. As shown in \autoref{fig:cross_modalities}, the model successfully infers a bond masked by a black patch in the input image. To deduce this bond, the model refers to the text description and applies chemistry rules. Specifically, if B'' is an oxygen atom O, there must be two connections to the rest of the molecule, indicating a double bond. Conversely, if B'' is an alcohol group OH, there only must be a single connection, resulting in a single bond. This ability to understand cross-modal dependencies is enabled by our joint visual and textual recognition approach, which is helpful for accurately recognizing complex multi-modal Markush structures. (Cf. suppl. material for further analysis.)

\begin{table}[t]\vspace{-3mm}
\centering
\caption{\textbf{MarkushGrapher encoder analysis.} Comparison of encoders in MarkushGrapher: VTL only, VTL and OCSR with early fusion (EF) or late fusion (LF).}\vspace{-3mm}
\label{tab:encoder}
\resizebox{0.8\linewidth}{!}{
\begin{tabular}{llcclcc}
\hline
\multirow{2}{*}{\textbf{Methods}} &  & \multicolumn{2}{c}{M2S} &  & \multicolumn{2}{c}{USPTO-Markush} \\ \cline{3-4} \cline{6-7} 
 &  & EM & T &  & EM & T \\ \hline
VTL &  & \textbf{38} & \textbf{77} &  & 23 & 71 \\
VTL + OCSR (EF) &  & \textbf{38} & \textbf{77} &  & 28 & \textbf{76} \\
VTL + OCSR (LF) &  & \textbf{38} & 76 &  & \textbf{32} & 74 \\ \hline
\end{tabular}} \vspace{-1mm}
\end{table}

\begin{table}[t]\vspace{0mm}
\centering
\caption{\textbf{Markush structure representation analysis.} Performance of MarkushGrapher using optimized CXSMILES (Optimized), optimized CXSMILES without the R-groups compression (Compression), and optimized CXSMILES without the atom indexing (Indexing). }\vspace{-3mm}
\label{tab:representation}
\resizebox{0.8\linewidth}{!}{
\begin{tabular}{llcclcc}
\hline
\multirow{2}{*}{\textbf{Methods}} &  & \multicolumn{2}{c}{M2S} &  & \multicolumn{2}{c}{USPTO-Markush} \\ \cline{3-4} \cline{6-7} 
 &  & EM & T &  & EM & T \\ \hline
Optimized &  & \textbf{38} & \textbf{76} &  & \textbf{32} & \textbf{74} \\
\hspace{3pt} - Compression &  & 30 & 68 &  & 31 & 70 \\
\hspace{3pt} - Indexing &  & 35 & 71 &  & 24 & 79 \\ \hline
\end{tabular}} \vspace{-4mm}
\end{table}

\section{Conclusion}
We propose a novel architecture to recognize multi-modal Markush structures in documents by combining a Vision-Text-Layout encoder and an Optical Chemical Structure Recognition encoder. Our model jointly process visual and textual definitions of a Markush structure and converts these to a structured graph and table. The model is trained on synthetic images and demonstrates strong generalization capabilities, allowing it to outperform existing methods on real-world data including our new M2S dataset. MarkushGrapher is a step towards large-scale extraction of Markush structures in documents, a key challenge in patent analysis \cite{SIMMONS2003195}.

\newcommand*{\dictchar}[1]{
    \clearpage
    \twocolumn[
    \centerline{\parbox[c][2.0cm][c]{18cm}{
            \centering
            \fontsize{15}{15}
            \selectfont
            {#1}}}]
}

\begin{center}
\dictchar{\vspace{-20mm}\textbf{Supplementary materials for \break MarkushGrapher: Joint Visual and Textual Recognition of Markush Structures\vspace{-10mm}}}
\end{center}

Here, we provide additional details and visualizations regarding Markush structures, the benchmark datasets, the MarkushGrapher analysis, and the synthetic training set, introduced in \autoref{section:Markush-Structure-Components}, \autoref{section:Benchmarks-Visual-Examples-and-Statistics}, \autoref{section:MarkushGrapher-Detailed-Analysis} and \autoref{section:Synthetic-Training-Set-Details}, respectively.

\section{Markush Structure}
\label{section:Markush-Structure-Components}

\begin{figure*}[b]
\centering
\resizebox{0.85\textwidth}{!}{
    \includegraphics*[trim={2cm 0cm 2cm 0cm}, clip, width=\textwidth]{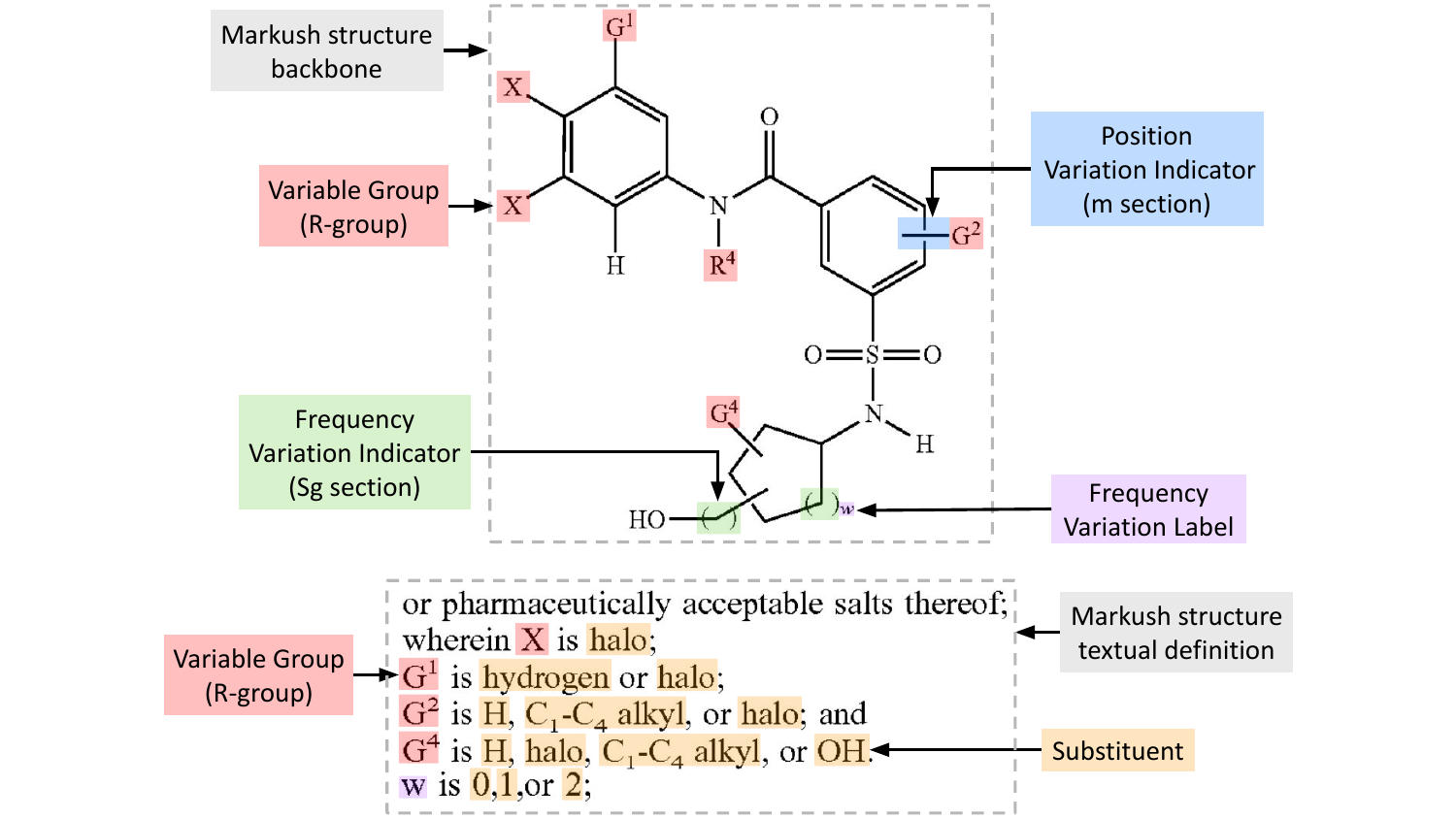}}\vspace{-2mm}
    \caption{\textbf{Markush structure components.}. Illustration of the two mains components of a Markush structure: the backbone and the textual definition. The backbone depicts the core of the chemical structure template: atoms, bonds, connectivity, variable groups (red), frequency variation indicators (green), and position variation indicators (blue). The textual definition lists substituents (orange) that can replace their respective variable groups and frequency variation labels in the backbone.}
    \label{fig:vocabulary}\vspace{0mm}
\end{figure*}

\autoref{fig:vocabulary} illustrates the different components of a Markush structure. A Markush structure contains two main components: a visual definition (referred to as Markush structure backbone) and a textual definition. The Markush structure backbone represents the core of the chemical structure template. It can identified with a Chemaxon Extended SMILES (CXSMILES) \cite{cxsmiles} string. For the Markush structure in \autoref{fig:vocabulary}, the CXSMILES is:

\begin{mybox}{CXSMILES}{}
[H]C1=C([*])C([*])=C([*])C=C1N(C)C(=O)C1=C
C=CC(=C1)S(=O)(=O)NC1CCCC1.CCO.*[*].*[*]\ \textbar\textcolor{darkred}{\$;;;X;;X;;G1;;;;;;;;;;;;;;;;;;;;;;;;;;G2;;G4\$},\textcolor{darkblue}{m:29:24.25.
26.27.28,m:32:14.19.15.18.17.16,m:34:24.25.26.27.
28},\textcolor{darkgreen}{Sg:n:28:w:ht,Sg:n:30:\ :ht}\textbar
\end{mybox}

\noindent The CXSMILES is composed of two sections. The first section holds the SMILES (in black) that identifies the atoms, the bonds and the connectivity of the structure. The second section is an extension table. It contains the variable groups (in red), the position variation indicators (sections starting with `m', in blue) and the frequency variation indicators (sections starting with `Sg', in green). The numbers written in the position variation and frequency variation indicators correspond to the index of the atoms in the SMILES. More details can be found in the CXSMILES documentation \cite{cxsmiles}. As shown in \autoref{fig:vocabulary}, the textual definition of the Markush structure defines the possible substituents for the different variable groups and frequency variation labels depicted in the Markush structure backbone. 

\section{Benchmark Datasets}
\label{section:Benchmarks-Visual-Examples-and-Statistics}

\subsection{Document Selection}
To build M2S, we manually sample documents published by the US Patent and Trademark Office (USPTO), European Patent Office (EPO) and World Intellectual Property Organization (WIPO). The selected patents are published between 1999 and 2023. 

To build USPTO-Markush, we sample images published by the USPTO between 2010 and 2016.

\subsection{Visual Examples}

\begin{figure*}[]
\centering
\resizebox{\textwidth}{!}{
    \includegraphics*[trim={13.5cm 0cm 13.5cm 0cm}, clip, width=\textwidth]{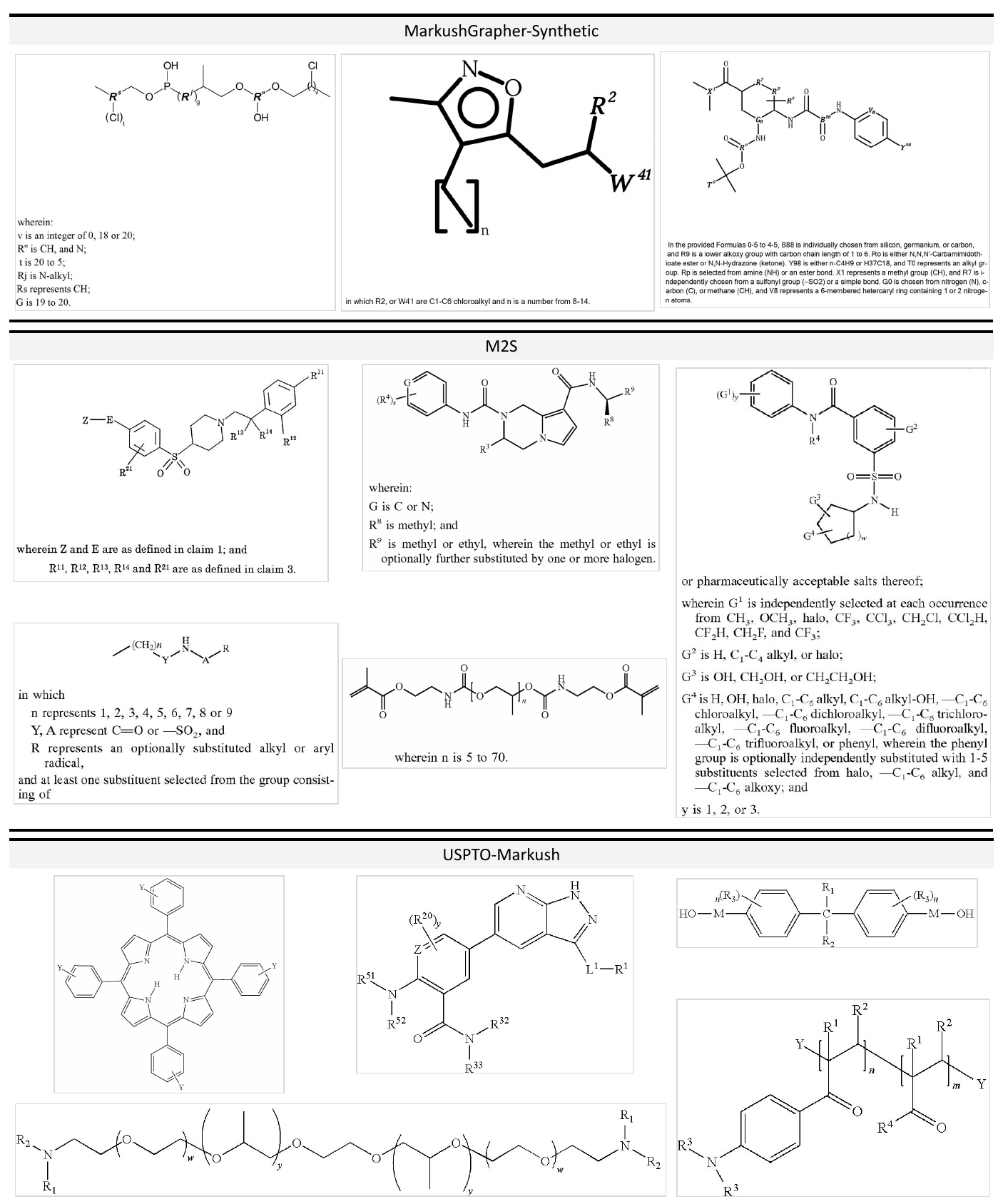}}
    \caption{\textbf{Benchmarks example images.} Example images randomly selected in MarkushGrapher-Synthetic, M2S, and USPTO-Markush.}
    \label{fig:benchmarks}
\end{figure*}

\autoref{fig:benchmarks} illustrates some images randomly sampled from MarkushGrapher-Synthetic, M2S and USPTO-Markush.

\begin{table*}[b]\vspace{0mm}
\centering
\caption{\textbf{Benchmarks statistics.} Comparison of the number of samples, the proportion of images containing each Markush structure features (R-group, `m' section, `Sg' section), the number of atoms, the number of variable groups and frequency variation labels, and the number of substituents for the different benchmarks.}\vspace{-3mm}
\label{tab:statistics}
\resizebox{\linewidth}{!}{
\begin{tabular}{lclccclccc}
\hline
\multirow{4}{*}{\textbf{Benchmarks}} & \multirow{4}{*}{\begin{tabular}[c]{@{}c@{}}Number of \\ samples\end{tabular}} &  & \multicolumn{3}{c}{\multirow{2}{*}{\begin{tabular}[c]{@{}c@{}}Proportion of CXSMILES\\ with at least one:\end{tabular}}} &  & \multirow{4}{*}{\begin{tabular}[c]{@{}c@{}}Mean number \\ of atoms\end{tabular}} & \multirow{4}{*}{\begin{tabular}[c]{@{}c@{}}Mean number of \\ variable groups \\ and frequency \\ variation label\end{tabular}} & \multirow{4}{*}{\begin{tabular}[c]{@{}c@{}}Mean number \\ of substituents\end{tabular}} \\
 &  &  & \multicolumn{3}{c}{} &  &  &  &  \\ \cline{4-6}
 &  &  & \multicolumn{1}{l}{\multirow{2}{*}{R-group}} & \multicolumn{1}{l}{\multirow{2}{*}{`m' section}} & \multicolumn{1}{l}{\multirow{2}{*}{`Sg' section}} &  &  &  &  \\
 &  &  & \multicolumn{1}{l}{} & \multicolumn{1}{l}{} & \multicolumn{1}{l}{} &  &  &  &  \\ \hline
MarkushGrapher-Synthetic & \textbf{1000} &  & 0.95 & 0.54 & 0.39 &  & \textbf{23} & 3.9 & \textbf{11} \\
M2S & 103 &  & \textbf{0.97} & 0.30 & 0.25 &  & 19 & 2.4 & 9.2 \\
USPTO-Markush & 74 &  & 0.91 & \textbf{0.74} & \textbf{0.42} &  & 20 & \textbf{4.7} & - \\ \hline
\end{tabular}} \vspace{0mm}
\end{table*}

\subsection{Statistics}

\autoref{tab:statistics} shows some statistics on MarkushGrapher-Synthetic, M2S and USPTO-Markush benchmarks. We observe that the three benchmarks contain a large fraction of Markush structures having R-groups. USPTO-Markush contains about twice as much images with `m' and `Sg' sections than M2S. 
Given that MolScribe is unable to predict `m' and `Sg' sections, this clarifies why MolScribe \cite{Qian2023} performs worse on USPTO-Markush than on M2S (see Table 1 of the main paper). Besides, the ablation study shown in Table 4 of the main paper demonstrates that adding the atom indices in the CXSMILES improves MarkushGrapher performance on USPTO-Markush substantially more than on M2S. It suggests that the atom indexing is particularly useful for predicting the `m' and `Sg' sections. 
\autoref{tab:statistics} also reports the mean number of atoms per sample, reflecting the average size of Markush structure backbones. It is similar for all three benchmarks.
Additionally, \autoref{tab:statistics} reports the mean number of variable groups, frequency variation labels, and substituents. These metrics are correlated with the average length of textual definitions of Markush structures. These definitions are on average longer for MarkushGrapher-Synthetic compared to M2S.

\section{MarkushGrapher Detailed Analysis}
\label{section:MarkushGrapher-Detailed-Analysis}

\subsection{Impact of Input Modalities}

\begin{figure*}[]
\centering
\resizebox{\textwidth}{!}{
    \includegraphics*[trim={14.5cm 2.5cm 14.5cm 2.5cm}, clip, width=\textwidth]{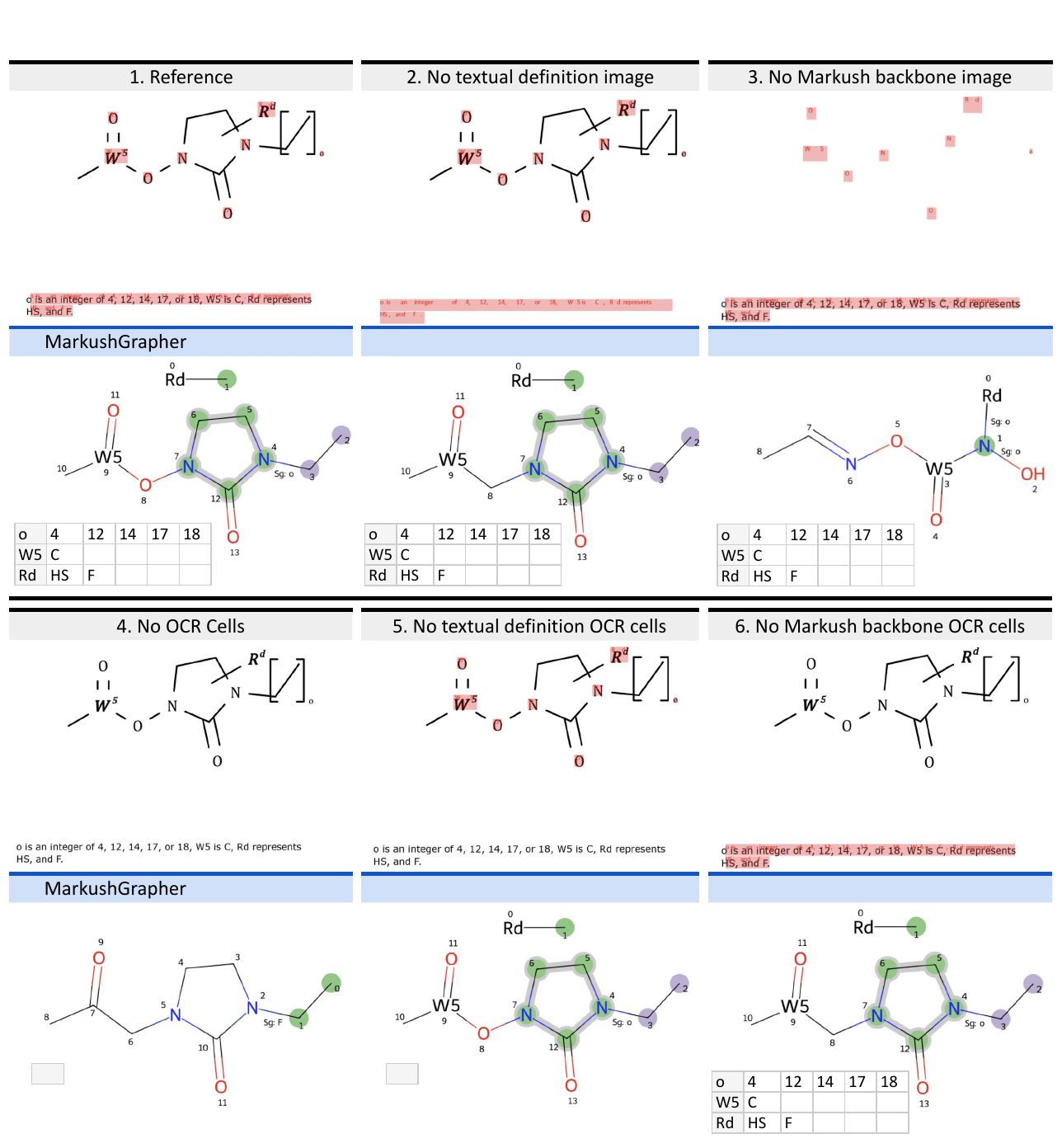}}\vspace{-4mm}
    \caption{\textbf{Modalities removal.} MarkushGrapher predictions after selectively removing components from a reference (input 1). The input OCR boxes are highlighted in red, with the corresponding OCR text written in red as well. In input 2, the Markush structure textual definition is removed from the image. In input 3, the Markush structure backbone is removed from the image. In input 4, all OCR cells are removed. In input 5, the OCR cells in the Markush structure textual definition are removed. In input 6, the OCR cells in the Markush structure backbone are removed.}
    \label{fig:modalities_impact}\vspace{20mm}
\end{figure*}

\begin{figure*}[]
\vspace{0mm}
\centering
\resizebox{\textwidth}{!}{
    \includegraphics*[trim={14.5cm 8.5cm 14.5cm 8.5cm}, clip, width=\textwidth]{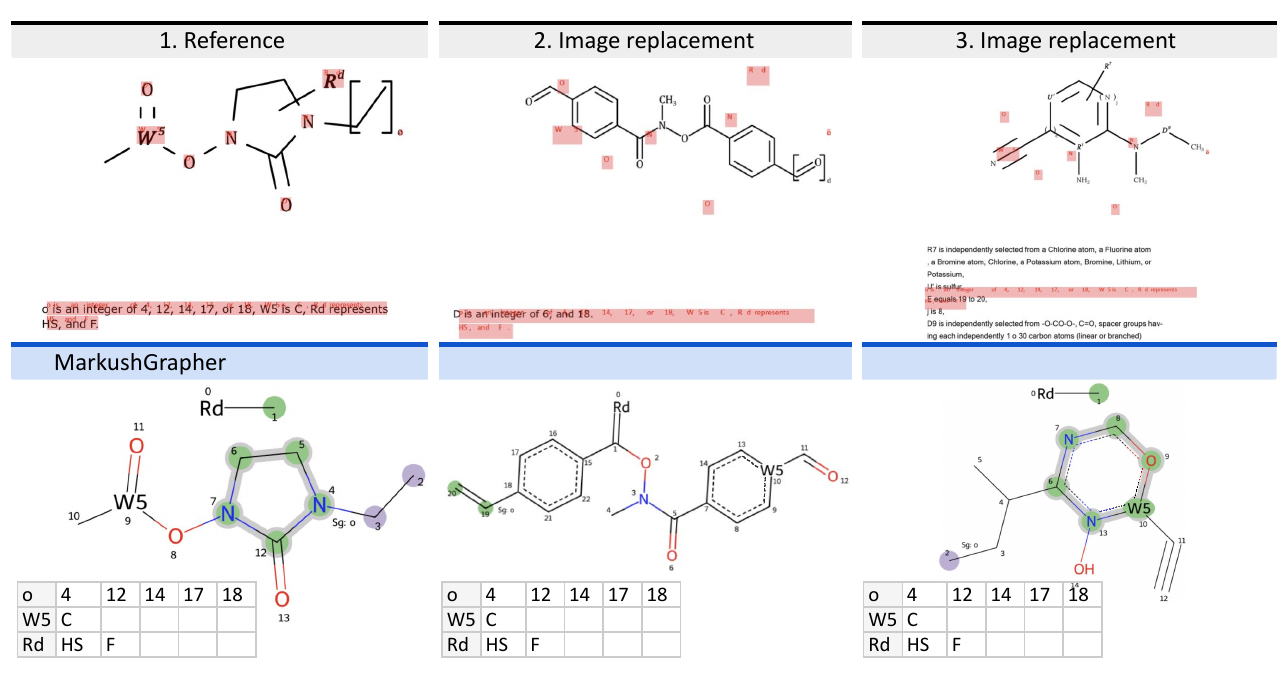}}\vspace{-4mm}
    \caption{\textbf{Image modality replacement.}. Examples of MarkushGrapher predictions after replacing input image 1 with either image input 2 or image input 3. The input OCR boxes are highlighted in red, with the corresponding OCR text written in red as well. The OCR cells remain unchanged.}
    \label{fig:modalities_impact_2}\vspace{-4mm}
\end{figure*}

Here, we analyze how each input modality contributes to the MarkushGrapher predictions.

\autoref{fig:modalities_impact} illustrates MarkushGrapher predictions after selectively removing components from an example input. We remove, one at a time, the Markush structure textual definition from the image, the Markush structure backbone from the image, all OCR cells, OCR cells from the Markush structure textual definition, and OCR cells from the Markush structure backbone. 
Inputs 2, 4 and 5 in \autoref{fig:modalities_impact} suggest that MarkushGrapher predicts the substituents tables using only OCR cells from the textual definition. The textual definition image appears unnecessary.
According to input 4 in \autoref{fig:modalities_impact}, MarkushGrapher appears to utilize the image of the Markush structure backbone to predict an initial structure. This first prediction represents the shape of the structure, \ie if all atom types and Markush structure features were ignored, the prediction would correct. Some common atoms such as oxygen or nitrogen are occasionally added to this initial structure, while the model does not have access to the OCR text provided during training. At this stage, most Markush structure features are ignored (R-groups and `m' sections) or incomplete (`Sg' sections).
Input 6 in \autoref{fig:modalities_impact} indicates that MarkushGrapher leverages OCR cells of the textual definition to know which variable group and frequency variation indicator need to be added to the initial structure.
Futhermore, input 3 in \autoref{fig:modalities_impact} is an indication that if no input backbone image is provided, MarkushGrapher attempts to infer a compact structure that connects the OCR cells. In this case, said structure respects the valence constraints of variable groups given by the textual definition.

\begin{figure*}[]
    \centering \vspace{-4mm}
    \includegraphics*[trim={14.5cm 1.5cm 14.5cm 1.6cm}, clip, width=0.875\textwidth]{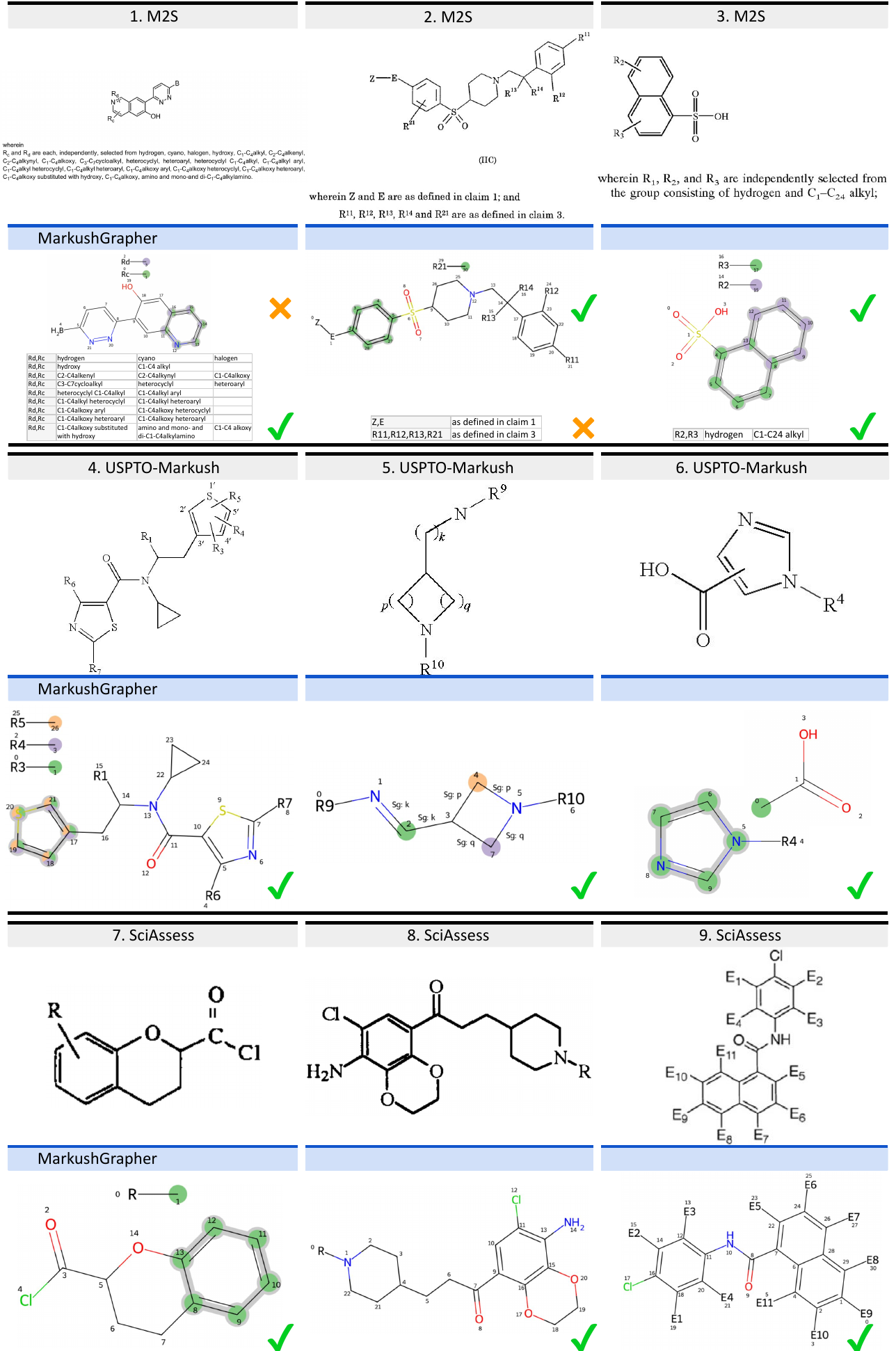}\vspace{-2mm}
    \caption{\textbf{Qualitative evaluation.} Examples of MarkushGrapher predictions are shown on real-world data: M2S (inputs 1, 2 and 3), USPTO-Markush (inputs 4, 5 and 6), SciAssess (inputs 7, 8 and 9).}
    \label{fig:qualitative}\vspace{0mm}
\end{figure*}

\begin{figure*}[]
\centering
\resizebox{\textwidth}{!}{
    \includegraphics*[trim={14.5cm 2.5cm 14.5cm 2.5cm}, clip, width=\textwidth]{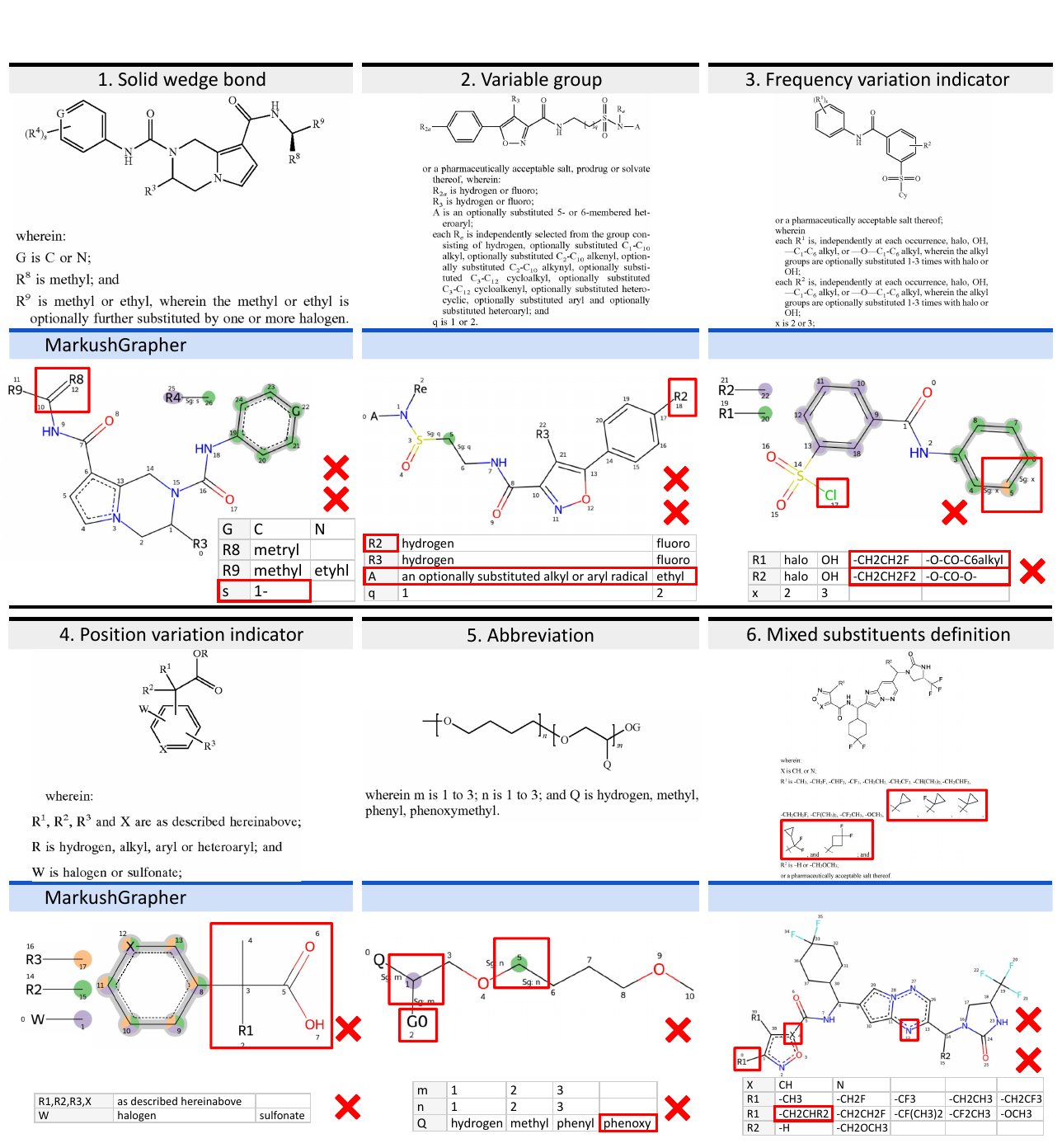}}\vspace{0mm}
    \caption{\textbf{Failure cases.} Example of failure cases of MarkushGrapher on real-world data from M2S. The errors are highlighted with red boxes. Typical failure cases include solid wedge bond (input 1), incorrect variable group label (input 2), incorrect frequency variation indicator (inputs 3 and 4), incorrect position variation indicator (input 4), unsupported abbreviation (input 5) and unsupported substituents definition (input 6).}
    \label{fig:fails}\vspace{15mm}
\end{figure*}

\autoref{fig:modalities_impact_2} shows examples of MarkushGrapher predictions after replacing the input image with alternative images while keeping the OCR cells unchanged. The inputs 2 and 3 in \autoref{fig:modalities_impact_2} seem to confirm that MarkushGrapher uses the image of the Markush structure backbone to predict the shape of the structure. The model appears to detect that some regions of this initial structure need to be replaced using the content of OCR cells. Most of the time, this replacement is then done using the closest OCR cell in the image. For example in input 2, the variable group `W5' is placed at the closest location where it has 4 connections, thus correctly respecting the valence constraints given by the Markush structure textual definition.

\subsection{Qualitative Evaluation}

\begin{figure*}[b]
\centering
\resizebox{\textwidth}{!}{
    \includegraphics*[trim={4cm 1cm 4cm 1cm}, clip, width=\textwidth]{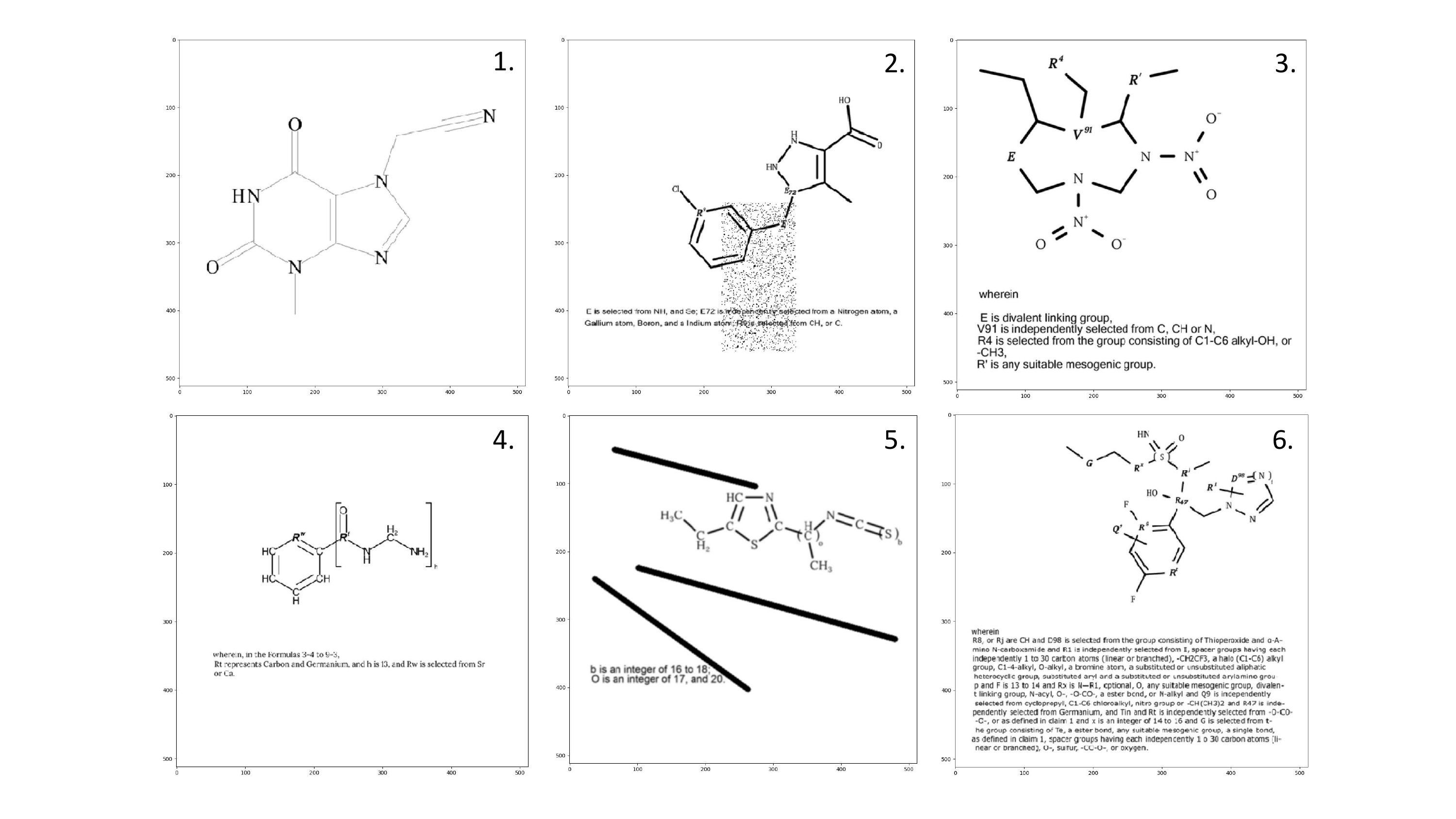}}\vspace{-2mm}
    \caption{\textbf{Examples of training images.} Randomly selected training images augmented by applying shifting (used in examples 1 to 6), scaling (used in examples 1 to 6), downscaling (used in examples 1 to 6), gaussian blur (used in examples 1 to 6), adding random pepper patches (used in example 2) and random lines (used in example 5).}
    \label{fig:training}\vspace{0mm}
\end{figure*}

\textbf{Qualitative examples.} \autoref{fig:qualitative} provides examples of MarkushGrapher predictions on real-world data. Notably, MarkushGrapher correctly predicts long tables (see input 1 in \autoref{fig:qualitative}). MarkushGrapher can also correctly recognize multi-modal Markush structures with drawing styles from the different patent offices. (In \autoref{fig:qualitative}, input 1 is published by EPO, input 2 is published by WIPO, and input 3 is published by USPTO.) MarkushGrapher also handle Markush structures backbones having a large number of variable groups (see input 9 in \autoref{fig:qualitative}), a large number of position variation indicators on the same cycle (see input 4 in \autoref{fig:qualitative}), a large number of frequency variation indicators (see input 5 in \autoref{fig:qualitative}), and function groups connected to cycles (see input 6 in \autoref{fig:qualitative}). Additionally, the model recognizes Markush backbones even when atom indices are displayed (see input 4 in \autoref{fig:qualitative}).

\noindent\textbf{Failure cases.} \autoref{fig:fails} illustrates examples of failure cases of MarkushGrapher on the M2S benchmark. Input 1 in \autoref{fig:fails} shows an inversion between a solid wedge bond and a double bond. Input 2 in \autoref{fig:fails} shows an incorrect variable group label. While the OCR text correctly contains `R2a', MarkushGrapher only predicts `R2'. It is probably due to the OCR cells augmentations used during training (see \autoref{section:Synthetic-Training-Set-Details}) and the abundance of `R2' as a variable group label. Input 3 in \autoref{fig:fails} highlights an incorrect prediction for a frequency variation indicator. In this example, the bounding box of the frequency variation indicator label is near to a carbon atom, and then incorrectly associated with this atom. Similary, input 4 in \autoref{fig:fails} presents an incorrect prediction for a position variation indicator. Here, a complex substructure is attached to a cycle. The model struggles with this because, during training, it only encounters R-groups and functional groups connected to cycles. The input 5 in \autoref{fig:fails} shows an incorrect abbreviation prediction. The input image contains the abbreviation `OG' (oxygen atom connected to variable group G) but MarkushGrapher predicts a variable group `G0', as it does not currently support abbreviations. On the same image, the frequency variation indicators represented with brackets are also only partially predicted. Input 6 in \autoref{fig:fails} shows a substituent definition that combines text and interleaved chemical structure drawings. This challenging setup is currently not supported by MarkushGrapher.
Additionally, we observe that the predicted substituent table occasionally contain additional labels, which are not in the input (see input 1 in \autoref{fig:fails}), as well as missing labels (see input 4 in \autoref{fig:fails}). 

Besides, It is worth noting that MarkushGrapher is trained to predict `m' sections which connect to all atoms in a cycle. Strictly speaking, this could be seen as incorrect, as some connections violate valence constraints. However, the MarkushGrapher output contains all information needed to reconstruct only the valid connections.

\begin{table}[t]\vspace{0mm}
\centering
\caption{\textbf{MarkushGrapher robustness.} Exact match accuracy is reported on augmented (A) versions of the real-world benchmark.}\vspace{0mm}
\label{tab:robustness}
\resizebox{\linewidth}{!}{
\begin{tabular}{lllll}
\hline
Method         & \multicolumn{2}{c}{M2S-100-A}                              &  & \multicolumn{1}{c}{USPTO-Markush-A} \\ \cline{2-3} \cline{5-5} 
               & \multicolumn{1}{c}{CXSMILES} & \multicolumn{1}{c}{Table} &  & \multicolumn{1}{c}{CXSMILES}      \\ \hline
MarkushGrapher &         \hspace{6mm}31                     &       \hspace{2mm}28                    &  &              \hspace{11mm}32                     \\ \hline
\end{tabular}} \vspace{-3mm}
\end{table}

\subsection{Model robustness}

Despite being trained on synthetic data, MarkushGrapher generalizes well to real-world datasets like M2S and USPTO-M. To further validate this, we tested the model on augmented versions of these benchmarks, simulating low-quality inputs such as scanned documents, using the same augmentations applied during training (see examples of augmentations in \autoref{fig:training}). \autoref{tab:robustness} shows that MarkushGrapher maintains strong performance in these challenging scenarios.

\subsection{Inference on real-data}

MarkushGrapher currently relies on ground-truth OCR cells as input. To enhance usability, OCR cells should be obtained through an OCR model. One possible approach is to use the abbreviation recognition approach introduced in MolGrapher \cite{Morin_2023_ICCV}.
In this method, text cells are extracted from chemical images using rule-based processing and PaddleOCR \cite{PaddleOCR}. Candidate text positions are identified by filtering connected components based on size, followed by character detection and recognition with PaddleOCR. Post-processing then corrects common chemical symbol inversions.
Using this approach, we annotated 100 images from the standard USPTO benchmark. Manual inspection showed that 95\% of predicted text cells were correct. To further improve OCR quality, training a dedicated OCR model for Markush structure images would be necessary.

\section{Synthetic Training Set Details}
\label{section:Synthetic-Training-Set-Details}

\subsection{Visualization and Image Augmentation}

\autoref{fig:training} shows randomly selected training images. A small portion of training images are standard chemical structure images (see example 1 in \autoref{fig:training}). Some training samples contain short (see example 2 in \autoref{fig:training}) or long (see example 6 in \autoref{fig:training}) textual definitions. 
Training images are generated using synthetic CXSMILES. To create them, we use SMILES from the PubChem \cite{10.1093/nar/gky1033} database and augment them using the RDKit \cite{RDKit} library. Based on predefined probabilities, we randomly:
\begin{itemize}
    \item Replace atom labels by variable groups (except for atoms with charges),
    \item Add parentheses on atoms,
    \item Add brackets on pairs of atoms (except for atoms in rings), 
    \item Connect [R-*] fragments to atoms in rings,
    \item Connect [R-*] fragments to rings,
    \item Connect functional groups to rings.
\end{itemize}
RDKit’s sanitization ensures that the generated structures are chemically valid.
Then, images are augmented by applying shifting, scaling, downscaling, gaussian blur, adding random pepper patches and random lines.
The generated structures are chemically correct but can be probably unlikely, due to the automatic generation of substituent definitions. For example, the image 6 in \autoref{fig:training} gives for the variable group `R47' the possible substituent `Germanium'. It would be chemically unlikely given to the rest of the molecule.

\subsection{Textual Definition Augmentation}

Markush structure textual definition are generated using manually-created templates.
A fraction of these definitions is then paraphrased with Mistral-7B-Instruct-v0.3 \cite{jiang2024mixtralexperts}, using the prompt:

\begin{mybox}{Prompt}{}
I want you to augment a text description. Paraphrase it without changing its semantic meaning, but only its formulation. Do not add or remove any information. Use the writing style of patents in the chemistry domain. To help you preserving the semantic meaning of the description, a dictionary is also provided. Its keys and values should not be modified in the augmented text description. Directly answer with one augmented text description, and nothing else. Do not give any dictionary output. Text description (to be paraphrased): \textbf{\textcolor{darkred}{Description}}. Dictionary input (for context only): \textbf{\textcolor{darkgreen}{Substituent table}}.
\end{mybox}

\noindent The variables `description' and `substituent\_table' are replaced for each textual definition to be augmented.
For example, if the initial description is:

\begin{mybox}{\textcolor{darkred}{Description}}{}
in which                                                                                          
M31 or Rj are lower alkyl, an alkenyl, CH(CH3)2, and heteroarylcarbonyl, and Ry is selected from the group consisting of N-alkyl, oxygen, a hydrocarbon group or optional, and Rp is selected from CH, C or N, and R4 is selected from a hydrocarbon group and N-aryl, and T9 represents C2-C6-alkenylcarbamoyl, and M4 represents a Nitrogen atom, and E is selected from the group consisting of NH, sulfur, hydrocarbon group or -O-CO-, and W1 is a Beryllium atom.  
\end{mybox}

\noindent And the substituent table is:

\begin{mybox}{\textcolor{darkgreen}{Substituent table}}{}
\{

\hspace{10pt}'M31': [

\hspace{10pt}\hspace{10pt}'lower alkyl', 'an alkenyl', 'CH(CH3)2', 

\hspace{10pt}\hspace{10pt}'heteroarylcarbonyl'

\hspace{10pt}], 

\hspace{10pt}'Rj': [

\hspace{10pt}\hspace{10pt}'lower alkyl', 'an alkenyl', 'CH(CH3)2', 

\hspace{10pt}\hspace{10pt}'heteroarylcarbonyl'

\hspace{10pt}], 

\hspace{10pt}'Ry': [

\hspace{10pt}\hspace{10pt}'N-alkyl', 'oxygen', 'a hydrocarbon group', 

\hspace{10pt}\hspace{10pt}'optional'

\hspace{10pt}], 

\hspace{10pt}'Rp': [

\hspace{10pt}\hspace{10pt}'CH', 'C', 'N'

\hspace{10pt}], 
 
\hspace{10pt}'R4': [

\hspace{10pt}\hspace{10pt}'a hydrocarbon group', 'N-aryl'

\hspace{10pt}], 

\hspace{10pt}'T9': [

\hspace{10pt}\hspace{10pt}'C2-C6-alkenylcarbamoyl'

\hspace{10pt}], 

\hspace{10pt}'M4': [

\hspace{10pt}\hspace{10pt}'a Nitrogen atom'

\hspace{10pt}], 

\hspace{10pt}'E': [

\hspace{10pt}\hspace{10pt}'NH', 'sulfur', 'hydrocarbon group', '-O-CO-'

\hspace{10pt}],

\hspace{10pt}'W1': [

\hspace{10pt}\hspace{10pt}'a Beryllium atom'

\hspace{10pt}]

\} 
\end{mybox}

\noindent The resulting paraphrased description is:

\begin{mybox}{Paraphrased description}{}
In said composition, M31 or Rj may encompass lower alkyl, alkenyl, CH(CH3)2, and heteroarylcarbonyl, while Ry may be chosen from the group consisting of N-alkyl, oxygen, a hydrocarbon group, or an optional group. Rp may be selected from CH, C, or N, and R4 may be chosen from a hydrocarbon group or N-aryl. T9 represents C2-C6-alkenylcarbamoyl, and M4 represents a Nitrogen atom. E may be selected from the group consisting of NH, sulfur, a hydrocarbon group, or -O-CO-, and W1 denotes a Beryllium atom.
\end{mybox}

\subsection{OCR Cells Augmentation}

Using our synthetic generation pipeline, we can generate multi-modal Markush structure images along with OCR cells of all text in the image. During training, we apply augmentations to the OCR cells by shifting the position of OCR boxes, and modifying the OCR text by simulating OCR errors. These text augmentations include character substitution, character insertion, character deletion, characters transposition and case alteration.

\section{Limitations and Future Works}

Currently, we made the choice to not handle abbreviations in MarkushGrapher.
As future work, we aim to train an OCR dedicated to the detection and recognition of text in multi-modal Markush structures images. We plan to apply MarkushGrapher at scale to build a large scale database of Markush structures and make it searchable by extending Markush structures encoding techniques  \cite{doi:10.1021/ci00031a010,doi:10.1021/ci3000387}.

{
    \small
    \bibliographystyle{ieeenat_fullname}
    \bibliography{main}
}

\end{document}